# A cybersecurity AI agent selection and decision support framework

Masike Malatji

*Graduate School of Business Leadership (SBL), University of South Africa (UNISA)*
*Midrand, Johannesburg, South Africa, PO Box 392, Unisa, 0003*
malatm1@unisa.ac.za
https://orcid.org/0000-0002-9893-9598

**ABSTRACT**
This paper presents a novel, structured decision support framework that systematically aligns diverse artificial intelligence (AI) agent architectures—reactive, cognitive, hybrid, and learning—with the comprehensive National Institute of Standards and Technology (NIST) Cybersecurity Framework (CSF) 2.0. By integrating agent theory with industry guidelines, this framework provides a transparent and stepwise methodology for selecting and deploying AI solutions to address contemporary cyber threats. Employing a granular decomposition of NIST CSF 2.0 functions into specific tasks, the study links essential AI agent properties such as autonomy, adaptive learning, and real-time responsiveness to each subcategory's security requirements. In addition, it outlines graduated levels of autonomy (assisted, augmented, and fully autonomous) to accommodate organisations at varying stages of cybersecurity maturity. This holistic approach transcends isolated AI applications, providing a unified detection, incident response, and governance strategy. Through conceptual validation, the framework demonstrates how tailored AI agent deployments can align with real-world constraints and risk profiles, enhancing situational awareness, accelerating response times, and fortifying long-term resilience via adaptive risk management. Ultimately, this research bridges the gap between theoretical AI constructs and operational cybersecurity demands, establishing a foundation for robust, empirically validated multi-agent systems that adhere to industry standards.

# 1. Introduction

The integration of artificial intelligence (AI) has fundamentally reshaped the contemporary cybersecurity landscape, emerging as a critical driver in the ongoing battle against increasingly sophisticated cyber threats. While traditional security paradigms, reliant on rule-based or signature-based detection, struggle to keep pace with the dynamic and evolving nature of modern attacks – characterised by the proliferation of novel exploits, polymorphic malware, and zero-day vulnerabilities [1], [2] – AI offers a paradigm shift. Leveraging the power of machine learning (ML) and deep learning (DL) algorithms, AI demonstrates unparalleled proficiency in real-time analysis of vast datasets, enabling the identification of subtle and often elusive indicators of compromise [3], [4]. This advancement empowers AI-driven systems with potent techniques such as pattern recognition, anomaly detection, and predictive analytics, automating core cybersecurity functions like intrusion detection and proactive threat hunting [5], [6]. This automation minimises the reliance on manual intervention. It addresses the critical shortage of skilled cybersecurity professionals by providing invaluable support to Security Operations Centres (SOCs) in maintaining continuous surveillance and managing high-volume incident streams [7], [8].

However, the attributes that position AI as a formidable defensive tool – its inherent automation, rapid learning capabilities, and capacity for generating novel insights – also present new avenues for exploitation by malicious actors. This necessitates a proactive and adaptive approach to cybersecurity. The evolving landscape demands the strategic adoption of AI-based defensive measures and the development of resilient, explainable AI (XAI) models capable of dynamically adapting to evolving attack vectors while preserving transparency and fostering user trust [9], [10].

The true power of AI in cybersecurity lies in its synergistic combination with human expertise, enabling a proactive, scalable, and intelligent approach to threat management [11]. However, realising this potential necessitates a sound and multifaceted security strategy. This strategy must extend beyond mere technological implementation to encompass good governance frameworks, address critical ethical considerations, and prioritise thorough validation of AI model performance [12], [13]. A cornerstone of responsible AI integration in cybersecurity is its alignment with established industry frameworks, standards, and guidelines. The National Institute of Standards and Technology (NIST) Cybersecurity Framework (CSF), particularly its updated version 2.0 [14], provides a critical foundation for this alignment. Fig. 1 illustrates the NIST CSF 2.0, a versatile framework designed to empower organisations across diverse sectors, including government agencies, in effectively managing their cybersecurity risks [14], [15].

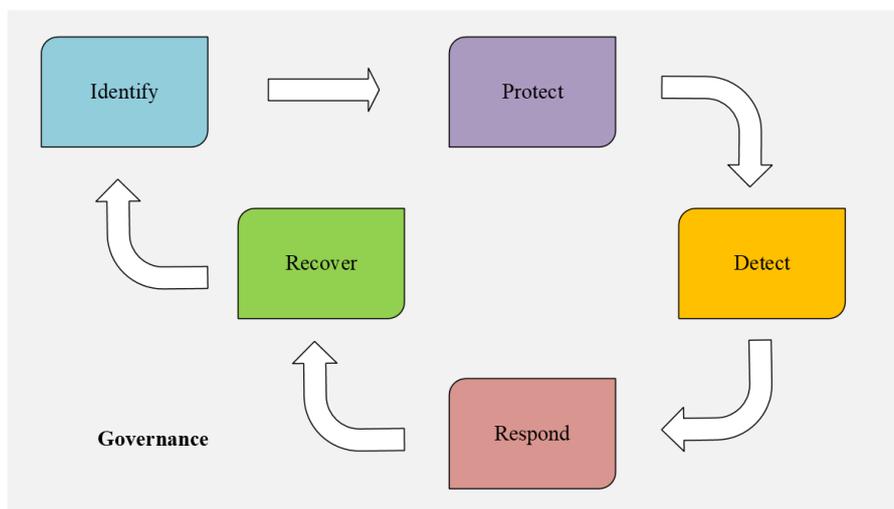

**Fig. 1** NIST CSF 2.0: Illustrates the high-level structure of the NIST Cybersecurity Framework 2.0, highlighting its six core functions and how they interrelate

The NIST CSF 2.0 offers a comprehensive taxonomy of high-level cybersecurity outcomes, serving as a valuable tool for organisations of all sizes, sectors, and maturity levels. It facilitates a more precise understanding, assessment, prioritisation, and communication of cybersecurity efforts [14]. Notably, the NIST CSF 2.0 is not prescriptive in its implementation; it serves as a guiding resource, complemented by supplementary guidance on potential practices and controls that can be leveraged to achieve the desired security outcomes [14]. As visually represented in Fig. 1, the NIST CSF 2.0 is organised around a cohesive set of six core functions that underpin a resilient cybersecurity posture: Identify, Protect, Detect, Respond, Recover, and Govern. These interconnected functions provide a comprehensive framework for managing and mitigating organisational cybersecurity risks. According to the NIST [14] guidelines, the core functions of the NIST CSF 2.0 are defined as follows:

- *Govern*: This overarching function establishes the importance of an organisation's commitment to defining, communicating, and consistently monitoring its cybersecurity risk management strategy, expectations, and policies.
- *Identify:* This foundational function emphasises the critical need for an organisation to understand its current cybersecurity risks thoroughly.
- *Protect:* This function focuses on implementing adequate safeguards to manage and mitigate identified cybersecurity risks.
- *Detect*: This involves the timely identification and analysis of potential cybersecurity attacks and compromises.
- *Respond*: This function outlines the necessary actions to address and minimise the impact of detected cybersecurity incidents.
- *Recover*: This focuses on the swift restoration of assets and operations affected by a cybersecurity incident to a functional state.

The practical implementation of the NIST CSF 2.0, however, can be significantly enhanced by leveraging the capabilities of advanced technologies. As mentioned earlier, AI has the potential to offer powerful tools to operationalise and strengthen these core cybersecurity functions. Indeed, as the sophistication of cyber threats continues to escalate [16], the need for comprehensive, adaptive, and scalable solutions across heterogeneous digital ecosystems becomes increasingly critical [17], [18]. With their inherent capacity for autonomous reasoning, real-time decision-making, and collaborative learning [19], [20], AI agents have emerged as compelling candidates to meet these evolving demands. AI agents are computational entities, often comprising software and sometimes hardware components, designed to exhibit autonomy, social interaction, responsiveness, and proactive behaviour [21], [22], [23], [24]. These agents can perceive their environment and execute actions to achieve predefined objectives [25]. While definitions may vary across disciplines, a common thread emphasises their ability to function independently, adapt to dynamic conditions, and interact with other agents or systems [24], [26], [27]. Within the cybersecurity domain, AI agents hold significant potential for automating critical tasks, such as threat monitoring, detection, and incident response, by effectively leveraging both reactive and proactive decision-making strategies.

From an architectural standpoint, the landscape of AI agents encompasses a diverse range of designs, each tailored to specific functionalities and environments. Key categories include virtual agents, which operate within digital environments; embodied agents, which possess a physical presence and interact with the physical world; reactive agents, characterised by their immediate response to environmental stimuli; hybrid agents, integrating multiple architectural paradigms; learning agents, which possess the capacity to adapt and improve their performance over time; and cognitive or deliberative agents, characterised by higher-level functions such as planning, reasoning, and learning [21], [28], [29], [30], [31], [32]. Categorising AI agents provides a framework for researchers and practitioners to strategically align agent capabilities and design limitations with the specific requirements of diverse application domains. This is particularly useful in modern cybersecurity, where layered defence and dynamic threat response strategies demand tailored technological solutions. To this end, an appropriate, theory-driven mapping of AI agent architectures to the NIST CSF 2.0 is required. Such a mapping ensures that the technical strengths of various agent types are effectively leveraged to support the core functions that underpin an organisation's cybersecurity posture.

However, the optimal AI agent architecture is not a one-size-fits-all solution. For instance, while reactive agents demonstrate exceptional proficiency in rapid, event-driven responses [29], they may lack the strategic foresight essential for proactive threat hunting or optimal resource allocation. Conversely, learning and cognitive agents offer sound planning and adaptive capabilities [28], but often come with increased computational and maintenance complexities [33]. Thus, research gaps persist in the current literature regarding the systematic and theoretically grounded integration of AI agents within established cybersecurity frameworks. In other words, while the potential of AI agents in bolstering cyber defences has been widely acknowledged, a cohesive and standardised framework for their implementation remains largely absent. A study by Kott [34] envisioned a future in which intelligent, autonomous cyber defence agents would play a pivotal role in threat mitigation. Subsequent research, including contributions from Théron and Kott [35] and Kott et al. [36], has further advanced this vision by proposing architectural models for autonomous intelligent cyber defence agents capable of outperforming human response speed and agility. Expanding the scope, Truong et al. [37] provided a broader perspective on the applications of AI agents across both offensive and defensive cybersecurity domains. Ligo et al. [38] delved into the intricacies of measuring cyber-resilience in systems that incorporate autonomous agents, highlighting the complexities of evaluating their effectiveness. Furthermore, Naik et al. [39] offered a review of AI techniques, including the role of AI agents in analysing, detecting, and mitigating diverse cyber threats. More recently, Sharma and Jindal [40] explored the broader implications of AI, underscoring its potential to advance intelligent agent technologies across various domains, including cybersecurity. Despite these valuable contributions, a unified, framework-driven approach to integrating and evaluating AI agents within established cybersecurity standards remains an area for further investigation.

In the evolving landscape of AI in cybersecurity, this paper aims to develop a framework for selecting, designing, and deploying AI agents to address the NIST CSF 2.0 cybersecurity requirements. The contribution of this framework is that it provides a structured lens through which to understand, for instance, how the adaptive learning capabilities of learning agents can enhance the Detect and Respond functions of the NIST CSF 2.0 by proactively adapting to emerging threat vectors [14], [41], [42]. Similarly, it clarifies how the strategic planning inherent in cognitive/deliberative agents can better facilitate the Governance function by effectively balancing compliance requirements with actionable threat intelligence [14], [43]. Without a clearly defined mapping, organisations risk deploying AI agents ill-suited to their specific security needs or failing to fully capitalise on the unique strengths of agents optimised for particular operational contexts. Therefore, rigorous conceptual mapping is vital for informing decision-makers and system architects, guiding them towards the most appropriate agent solutions tailored to their cybersecurity objectives. This alignment between conceptual rigour and practical applicability paves the way for empirical validation and targeted tool implementations, ultimately fostering effective and sustainable AI-driven security operations.

This study is grounded in foundational concepts from agent-based AI, particularly the classification and behavioural properties of AI agents as applied to cybersecurity [21], [22], [24]. The framework builds upon four core agent architectures, reactive, cognitive, hybrid, and learning agents, each defined by its level of autonomy, decision-making mechanisms, and interaction with dynamic environments [21], [28], [29], [30], [31], [32]. As mentioned earlier, the paper also draws from recognised cybersecurity governance models, primarily the NIST CSF 2.0, which serves as the anchor for mapping agent capabilities to structured cybersecurity functions. Additionally, the concept of graduated levels of autonomy is introduced, informing the selection logic that guides agent deployment across the Identify, Protect, Detect, Respond, Recover, and Govern functions of the NIST CSF 2.0. This layered autonomy spectrum, ranging from assisted [44], [45], augmented [46], [47], to fully autonomous intelligence [48], [49], [50], enables a capability-based interpretation of agent suitability aligned with organisational maturity, task complexity, and risk appetite. These theoretical constructs form the foundational lens through which the AI Agent Taxonomy and Decision Framework (AIATDF) is developed and validated in this paper.

The methodology for developing the framework employs a matrix-based approach to map specific agent properties (e.g., autonomy, adaptiveness, reactivity) to the subcategories and tasks defined by the NIST CSF 2.0. This mapping forms the basis for identifying which agent architectures (e.g., reactive, hybrid, learning) are best suited to different cybersecurity functions (e.g., Detect, Respond, Recover) [14], [28], [29]. In a nutshell, the proposed approach is grounded in a structured, matrix-based conceptual mapping of agent capabilities to cybersecurity requirements. Below is a concise overview of the *main contributions* presented in this paper:

- *AI Agent Taxonomy and Decision Framework (AIATDF):*
  Presents a structured model for classifying AI agents and systematically mapping their properties to the NIST CSF 2.0 functions (Fig. 6).
- *Conceptual mapping matrix:*
  Provides a detailed mapping matrix, linking specific AI agent types and capabilities to NIST CSF 2.0 subcategories, facilitating optimal architecture selection for real-world cybersecurity tasks (Table 6).
- *Graduated levels of autonomy:*
  Establishes a preliminary capability maturity model (CMM), enabling organisations with varying maturity levels to incrementally adopt AI-driven solutions, progressing from assisted to augmented and autonomous intelligence (Fig. 3).
- *Holistic NIST CSF 2.0 alignment:*
  Demonstrates the integrated deployment of AI agents across all NIST CSF 2.0 functions, providing an end-to-end cybersecurity management approach (Table 6).
- *Rigorous methodology and applied utility:*
  Combines hierarchical task analysis with AI agent theory, yielding a transparent and reproducible framework suitable for academic research and applied SOC deployments (Steps beneath Fig. 6).

The theoretical framework presented here is based on several core assumptions. Firstly, I assume that the NIST CSF 2.0 provides sufficiently granular and actionable functions and subcategories for mapping AI agent tasks in operational settings. Secondly, I contend that the inherent properties of AI agents, such as autonomy, reactivity, and learning capabilities, can be sufficiently characterised and conceptually aligned with established cybersecurity requirements. Thirdly, the proposed framework assumes a baseline level of digital infrastructure and security maturity within deploying organisations. Despite these foundational assumptions, the study's primary limitation lies in its conceptual genesis; the AIATDF remains to be validated through empirical testing in live cybersecurity environments. Furthermore, while drawing extensively from scholarly and industry literature, my classification of AI agent architectures and properties may inherently simplify dynamically evolving, complex hybrid systems. These limitations are elaborated upon in Section 6.3.

Building upon the foundational understanding established in the preceding content, this paper proceeds as follows: Section 2 provides a detailed review of the relevant literature, explicitly exploring the capabilities of various AI agent architectures, the tenets of the NIST CSF 2.0, and the rationale for systematically mapping these agent capabilities to the framework's functional requirements. Subsequently, Section 3 outlines the methodological approach employed in this study to develop this mapping framework. The framework is presented in Section 4 and conceptually validated in Section 5. Discussions of the mapping framework, its implications, and inherent limitations are provided in Section 6. Finally, Section 7 concludes the paper by offering key takeaways and practical recommendations for future cybersecurity practices and suggesting promising avenues for further research in this evolving domain.

## 2. Related Works

### 2.1 NIST CSF 2.0

The NIST CSF 2.0 is an updated and comprehensive set of guidelines, standards, and best practices developed by the NIST to help organisations manage and reduce cybersecurity risks [14]. NIST is a non-regulatory federal agency that provides the scientific and technical foundation for innovation, economic growth, and public safety in the United States of America. The NIST CSF 2.0 is designed to be flexible and adaptable for organisations of all sizes, sectors, and cybersecurity maturity levels [14].

#### 2.1.1 The evolution of the NIST CSF

The NIST CSF has undergone a significant and necessary evolution from its initial release as version 1.0 in 2014 to the more comprehensive and globally relevant version 2.0. This progression mirrors the escalating complexity and scale of contemporary cybersecurity threats, necessitating more resilient and adaptable security strategies. While the foundational framework in version 1.0 established five core functions – Identify, Protect, Detect, Respond, and Recover [51], [52], [53] – version 2.0 represents a notable enhancement. A key advancement in version 2.0 is the introduction of a dedicated "Govern" function, which explicitly elevates the importance of cybersecurity within an organisation's overall governance structure [14], [53]. Therefore, the NIST CSF 2.0 is structured around six core functions, which are further elaborated through twenty-two categories representing specific cybersecurity outcomes and one hundred and six subcategories detailing more granular technical and management activities [14]. Fig. 2 provides a visual representation of this layered structure.

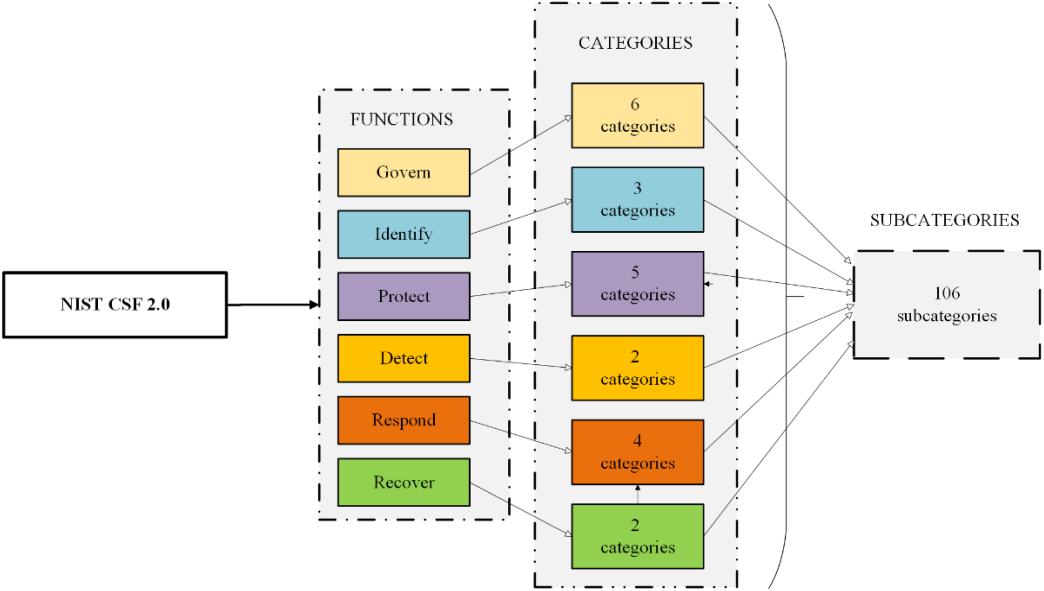

**Fig. 2** NIST CSF 2.0 framework core: Depicts the layered composition of the NIST Cybersecurity Framework 2.0, showing how the 22 categories and 106 subcategories are organised under the six core functions of the framework

The twenty-two NIST CSF 2.0 categories are summarised in Table 1 [14].

**Table 1** NIST CSF 2.0 categories: The 22 categories within the updated NIST Cybersecurity Framework 2.0 are listed, grouping them by the six core functions (Govern, Identify, Protect, Detect, Respond, Recover)

| Function | Category |
| --- | --- |
| Govern | o Organisational context |
| | o Risk management strategy |
| | o Roles, responsibilities, and authorities |
| | o Policy |
| | o Oversight |
| | o Cybersecurity supply chain risk management |
| Identify | o Asset management |
| | o Risk assessment |
| | o Improvement |
| Protect | o Identity management, authentication, and access control |
| | o Awareness and training |
| | o Data security |
| | o Platform security |
| | o Technology infrastructure resilience |
| Detect | o Continuous monitoring |
| | o Adverse event analysis |
| Respond | o Incident management |
| | o Incident analysis |
| | o Incident response reporting and communication |
| | o Incident mitigation |
| Recover | o Incident Recovery Plan Execution |
| | o Incident Recovery Communication |

Furthermore, the updated framework broadens its scope with a global perspective, ensuring its applicability and utility for worldwide organisations [52], [53]. Version 2.0 provides more detailed guidance and actionable recommendations, offering practical strategies and steps for cybersecurity professionals at all levels [53]. This enhanced version also explicitly addresses the rapid advancements in technology, including the Internet of Things, cloud computing, and big data, offering specific guidelines to manage the associated cybersecurity risks [14], [54], [55]. Moreover, the updated framework also emphasises a risk-based approach, equipping organisations with more solid methods for identifying, assessing, and effectively managing potential threats [14], [53]. Ultimately, version 2.0 empowers organisations to develop more comprehensive and threat-informed cybersecurity strategies tailored to the specific needs of diverse sectors and industries [52], [53]. The expanded functions and detailed guidance in version 2.0 provide cybersecurity professionals with a more precise and actionable roadmap for implementing effective security measures, including hundreds of immediately applicable recommendations [14], [53]. This enhanced international applicability positions version 2.0 as a valuable tool for standardising cybersecurity practices and fostering greater global cybersecurity resilience [14], [53].

However, because the functions, categories and subcategories are designed to be high-level and adaptable to various organisations, they only outline *what* cybersecurity outcomes should be achieved, but not necessarily *how*. 'Informative references' of the NIST CSF 2.0 provide the "how" by pointing to specific standards, guidelines, and best practices that offer detailed implementation guidance [14]. This is the level at which more granular cybersecurity controls are found and implemented. Thus, a clearly defined mapping framework is required to systematically examine the alignment between distinct AI agent capabilities and the functional pillars of the NIST CSF 2.0.

### 2.1.2 Rationale for mapping AI agents' capabilities to the NIST CSF 2.0

Reflecting the rapidly evolving cybersecurity landscape, NIST has proactively initiated an examination into how existing frameworks, such as the NIST CSF 2.0, can assist organisations in navigating emerging and expanding risks. As a testament to this, on February 14, 2025, NIST [56] published a concept paper for a cybersecurity and AI workshop, seeking public input on the critical challenge of addressing cybersecurity risks associated with the development and deployment of AI. The subsequent workshop, held on April 3, 2025, further underscored this focus. In essence, the NIST [56] concept paper centres on identifying and mitigating AI-related sources of cybersecurity risk that can significantly impact an organisation's operational risk profile. The concept paper categorises these risks into three key areas: (i) cybersecurity of AI systems, (ii) AI-enabled cyber attacks, and (iii) AI-enabled cyber defence, which align with the categories of adversarial AI, offensive AI, and defensive AI, respectively, previously described by Malatji and Tolah [57]. To quote NIST [56] directly, *"there is no consistent taxonomy or agreement on how AI advances inform organisations' strategies for cybersecurity risk management."*

They propose a "Cyber AI Profile" for guiding organisations deploying AI technologies and/or defending against AI-enabled attacks. At the time of writing, NIST is collaborating with its partners and stakeholders on the "Cyber AI Profile."

Against this backdrop, this paper's rationale for mapping AI agents to the NIST CSF 2.0 is that the strategic integration of these agents with the NIST CSF 2.0 offers a significant (autonomous) opportunity to fortify an organisation's cybersecurity posture across its core functions. This involves strategically leveraging AI's inherent capabilities to enhance asset identification, bolster threat detection, strengthen protective measures, streamline incident response, and accelerate recovery processes [4], [58], [59]. By embedding AI agents within these and other cybersecurity functions, organisations can anticipate even more improvements in response times and ensure more resilient, continuous protection against the ever-evolving threat landscape. Therefore, a structured mapping and decision-making tool is essential for effectively aligning the technical capabilities of diverse AI agent architectures with the specific security tasks that define an organisation's overall cybersecurity posture. Ultimately, the absence of such a structured mapping can directly contribute to security vulnerabilities and potentially lead to SOC underperformance due to the deployment of ill-suited AI agent solutions. In the next section, I review existing literature on the generic application of AI on the NIST CSF 2.0.

### 2.1.3    Application of AI on NIST CSF

This paper distinguishes between automation and autonomy within the context of AI/ML. From this perspective, the core differentiation hinges on the degree of intelligence embedded within the system and the extent of human involvement in the decision-making process [60], [61]. In this context, automation refers to AI-enhanced systems that execute predefined rules or algorithms to accomplish tasks while maintaining a significant degree of human control over the system's actions [62], [63]. These systems, even when employing sophisticated ML for optimisation, operate within the constraints of their initial programming. A typical example is a spam filter, which utilises algorithms and learned patterns to categorise emails based on pre-set parameters [64], demonstrating adaptability but not fundamentally altering its operational logic.

Conversely, autonomy describes AI systems characterised by their capacity for independent decision-making in dynamic and unpredictable environments [48], [49], [50]. These systems exhibit a greater degree of "agency" [65], enabling them to perceive their surroundings, interpret information, formulate plans, and execute actions to achieve objectives without explicit, step-by-step instructions for every possible scenario [24], [25], [26], [27]. A prime illustration is a self-driving car, which must navigate complex traffic scenarios, adapt to unexpected obstacles, and make real-time decisions regarding speed, direction, and braking [59], [66], [67]. While reliant on algorithms and pre-trained models, a self-driving car's ability to respond to novel and unforeseen situations distinguishes it from a purely automated system. Therefore, the fundamental distinction between automation and autonomy lies in the system's inherent capacity for adaptive behaviour, independent problem-solving, and the level of human intervention [48], [49], [50]. Automated systems excel in optimising predictable processes, whereas autonomous systems are designed to tackle unpredictable ones, necessitating continuous learning, adaptation, and real-time judgment [68]. This critical distinction underpins understanding the varying potential and limitations of different AI applications, a theme explored throughout this paper.

According to Morovat and Panda [69] and Dehghantanha et al. [68], cybersecurity paradigms evolve from traditional to AI-enhanced to autonomous. Traditional cybersecurity involves signature-based or rule-based defences, characterised by their reactive nature and limited capacity for dynamic analysis of novel threats [70]. This approach primarily responds to known threat patterns, lacking the agility to adapt to the ever-evolving adversary. AI-driven cybersecurity represents a significant advancement, integrating automated responses and leveraging ML for more flexible threat detection [10]. This shift transitions from static, rule-driven methodologies to sophisticated pattern-based analytics and accelerated incident triage. This paper focuses on the most advanced paradigm: autonomous cybersecurity. This paradigm leverages AI agents capable of reinforcement learning (RL), continuous self-improvement, and predictive orchestration across complex, distributed systems [71], [72], [73]. A core emphasis of this approach lies in developing XAI and implementing highly personalised threat defences that can rapidly adapt to the dynamic and unpredictable nature of modern attack landscapes [74], [75]. Furthermore, as AI agents are increasingly embedded in cybersecurity infrastructures, concerns about transparency, fairness, and privacy are becoming central to their design and deployment. As highlighted by Radanliev [76], ethical AI development demands proactive strategies that address data bias, algorithmic opacity, and differential privacy. These ethical safeguards are critical not only for user trust but also for regulatory compliance in sensitive domains like cybersecurity [9], [10], where decision-making must be both explainable and justifiable under audit conditions [77]. Incorporating such perspectives ensures that AI agents are not merely technically capable but also socially responsible and accountable.

Building upon the works of Dehghantanha et al. [68] and Morovat and Panda [69], this paper distinguishes four primary modes of AI system intelligence in cybersecurity. Fig. 3 illustrates the graduated levels of autonomy, ranging from purely manual (human-driven) to assisted intelligence, augmented intelligence, and ultimately, complete autonomy.

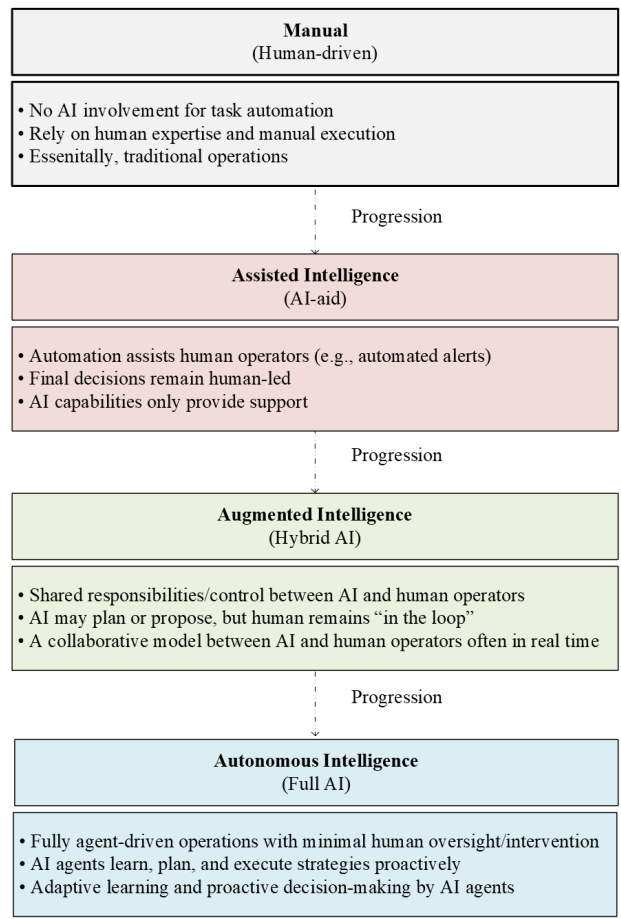

**Fig. 3** Graduated levels of autonomy in cybersecurity: Visualises a CMM's progression from manual, human-driven modes to assisted, augmented, and fully autonomous intelligence within cybersecurity contexts

The *manual* or human-driven mode in Fig. 3 represents a state without automation or AI integration, relying entirely on human expertise. *Assisted intelligence* enhances human capabilities and cognitive functions by providing information, automating specific tasks, or offering insights [44], [45]. Within this mode, AI systems act as tools to enhance human abilities for efficiency and effectiveness, with ultimate responsibility for outcomes remaining with human operators. A prime example is healthcare, where assistive AI aids clinicians in making more informed diagnostic or management decisions, as seen in applications ranging from virtual informatics systems for health management to physical robotic-assisted surgeries [78]. The previously mentioned cybersecurity spam filter also exemplifies this category, as the human operator retains responsibility for email filtering, positioning assisted intelligence as a form of AI/ML-driven automation. *Augmented intelligence* goes a step further than assisted intelligence. It is about a collaborative partnership between human intelligence and AI to amplify human capabilities by leveraging the unique strengths of both while mitigating their inherent weaknesses [46], [47].

While assisted intelligence inherently operates within the human-in-the-loop (HITL) human-machine interaction (HMI) paradigm, augmented intelligence possesses the capability to function across both HITL and the more advanced human-out-of-the-loop (full autonomy) HMI paradigm [79], [80], [81], [82], [83]. Therefore, augmented intelligence exhibits characteristics of automation and autonomy, effectively representing a hybrid mode that bridges assisted and autonomous intelligence. In cybersecurity, augmented intelligence enhances threat detection and response by combining human expertise with the analytical power of AI/ML, leading to more accurate and efficient handling of complex threats [84]. Finally, *autonomous intelligence*, as discussed at the beginning of this section, encompasses systems capable of independent task execution without direct human intervention, relying on advanced AI/ML techniques [48], [49], [50]. These systems are increasingly prevalent across diverse domains, including healthcare, military, transportation, and cybersecurity [83], [85], [86]. This paper focuses explicitly on AI agents operating as autonomous intelligent entities within the cybersecurity domain, while acknowledging their potential deployment in augmented intelligence mode. To the best of my knowledge, only Salesforce [87] released a similar AI agent CMM with graduated levels of autonomy, as shown in Fig. 3, on April 10, 2025. In other words, there is a shortage of scholarly, well-defined CMMs for the adoption of AI agents. This avenue forms part of my future research as it is beyond the scope of this paper.

Beyond the current paradigm of autonomous AI agents in cybersecurity, emerging research heralds a future of transformative capabilities. Frontier research explores quantum-safe AI algorithms to fortify defences against future quantum attacks [88] and privacy-preserving ML, including homomorphic encryption and differential privacy [89], to ensure data security without compromising agent efficacy. Federated learning is also emerging [90], promising decentralised AI pipelines that enable collaborative learning while preserving data locality and regulatory compliance. Though currently in development, these cutting-edge technologies hold the potential to redefine agent architectures and significantly enhance cybersecurity resilience. In this paper, thirty-five units of analysis were obtained from a systematic literature review (SLR) process through Scopus (12 papers), IEEE (5 papers), and Google Scholar (18 papers) databases. The key phrase *"AI agents in cybersecurity"* was utilised to search for the relevant literature. The notable outcomes of the SLR are summarised in Table 2.

**Table 2** Literature review of the application of AI on NIST CSF: Summarises existing studies that intersect cybersecurity, AI agents, and NIST CSF, indicating whether each study addressed the NIST CSF and its functions

| Author | Purpose of the study | Focus on cybersecurity | Focus on AI agents | Focus on NIST CSF |
|---|---|---|---|---|
| NIST [56] | Evaluate how an updated NIST CSF 2.0 can assist organisations facing new or expanded AI-related risks. | ✓ | ✗ | Potentially |
| Cronin [72] | Reviews generic AI agents and those that use Generative AI. | ✗ | ✓ | ✗ |
| Yu et al. [91] | Explore the potential for blockchain to be a foundational infrastructure for AI agents in the metaverse. | ✗ | ✓ | ✗ |
| Han et al. [92] | Propose a multi-agent system (MAS) to enhance financial investment research. | ✗ | ✓ | ✗ |
| Chan et al. [71] | Evaluate metrics to increase visibility into AI agents. | ✗ | ✓ | ✗ |
| Chen et al. [93] | Explore the role of 6G in realising AI agents' potential. | ✗ | ✓ | ✗ |
| Bovo et al. [94] | Explore large language models (LLM) agents in extended reality environments. | ✗ | ✓ | ✗ |
| Baabdullah [95] | Explores the impact of decision-making efficiency facilitated by generative conversational AI agents. | ✗ | ✓ | ✗ |
| Huang et al. [96] | Explore embodied AI agent systems. | ✗ | ✓ | ✗ |
| Agashe et al. [97] | Propose agent S, an open agentic framework. | ✗ | ✓ | ✗ |
| Kim and Saad [98] | Propose a novel continual learning algorithm for AI agents. | ✗ | ✓ | ✗ |
| Pleshakova et al. [99] | Create a neural network architecture for multi-agent tasks. | ✓ | ✓ | ✓ |
| Sharma and Jindal [40] | Explore the role of AI agents in various domains. | ✗ | ✓ | ✗ |
| Dehghantanha et al. [68] | Provide an overview of the current state of autonomous cybersecurity, highlighting the key challenges and opportunities. | ✓ | ✓ | ✗ |
| Hauptman et al. [100] | Explore how a team's work cycle could guide an AI agent's changing level of autonomy. | ✓ | ✓ | ✓ |
| Kott et al. [36] | Advance the reference architecture work on the AICA. | ✓ | ✓ | ✗ |
| Kaur et al. [101] | Analysed 236 AI use cases for cybersecurity provisioning against the NIST CSF 1.1. | ✓ | ✓ | ✓ |
| Cañas [102] | Studies human and AI agents' responsibility when interacting. | ✗ | ✓ | ✗ |
| Naik [39] | Review the application of AI techniques in fighting various cyberattacks. | ✓ | ✗ | ✗ |
| Roy [103] | Propose a MAS that detects and neutralises unseen cyber anomalies. | ✓ | ✓ | ✗ |
| Li et al. [104] | Train AI agents using the Cyber Gym for Intelligent Learning. | ✗ | ✓ | ✗ |
| Ligo et al. [38] | Examine approaches to measuring cyber-resilience systems with autonomous agents. | ✓ | ✓ | ✗ |
| Prasad et al. [77] | Evaluate AI agents' decisions using the Testing with Concept Activation Vectors XAI technique. | ✓ | ✓ | ✗ |
| Ashktorab et al. [105] | Investigate the social perceptions of AI agents. | ✗ | ✓ | ✗ |
| Zolotukhin et al. [106] | Explore attack mitigation techniques in software-defined networking (SDN) environments. | ✓ | ✓ | ✗ |
| Cao et al. [107] | Address fifth-generation (5G) SDN challenges with AI agents. | ✗ | ✓ | ✗ |
| Franco et al. [108] | Introduce a cybersecurity-driven conversational agent. | ✓ | ✓ | ✗ |
| Morovat and Panda [69] | Review the impact of AI on cybersecurity. | ✓ | ✗ | ✗ |
| Truong et al. [37] | Provide an overview of how AI can be used in cybersecurity for offensive and defensive AI. | ✓ | ✗ | ✗ |
| Kott and Théron [109] | Advance work on AICA and introduce its high-level reference architecture. | ✓ | ✓ | ✗ |
| Théron and Kott [35] | Advance work on AICA. | ✓ | ✓ | ✗ |
| Théron et al. [110] | Introduce the concept and architecture of an autonomous intelligent cyber-defence agent (AICA). | ✓ | ✓ | ✗ |
| Grzonka et al. [111] | Present a MAS cloud monitoring model. | ✗ | ✓ | ✗ |
| Yampolskiy [112] | Explains AI safety vs cybersecurity. | ✗ | ✓ | ✗ |
| Petrović [113] | Explores AI agents in virtual worlds. | ✗ | ✓ | ✗ |

A significant research gap is evident in the literature, as only three studies, indicated in Table 2, have examined the convergence of cybersecurity, AI agents, and the NIST CSF. Pleshakova et al. [99] briefly model attackers by neural networks, highlighting the capabilities of decentralised LLMs, or autonomous LLM agent swarms, into distinct cyber operations categories aligned with four functions of the NIST CSF 1.1 (Identify, Protect, Detect, Respond). Notwithstanding that the current paper concerns the NIST CSF 2.0, Pleshakova et al. [99] do not map any AI agent capabilities to the NIST CSF functions. Hauptman et al. [100] further narrowed the scope, concentrating solely on the Response function of the NIST CSF 1.1. Notably, Kaur et al. [101] recognised the importance of governance in cybersecurity preceding the release of NIST CSF 2.0. They astutely argued for AI's role in policy enforcement and risk monitoring, aligning with the framework's eventual inclusion of the 'Govern' function. However, as shown in Fig. 4, Kaur et al. [101] specification of AI-driven techniques was ultimately grounded in the outdated NIST CSF 1.1, missing critical updates and changes in NIST CSF 2.0.

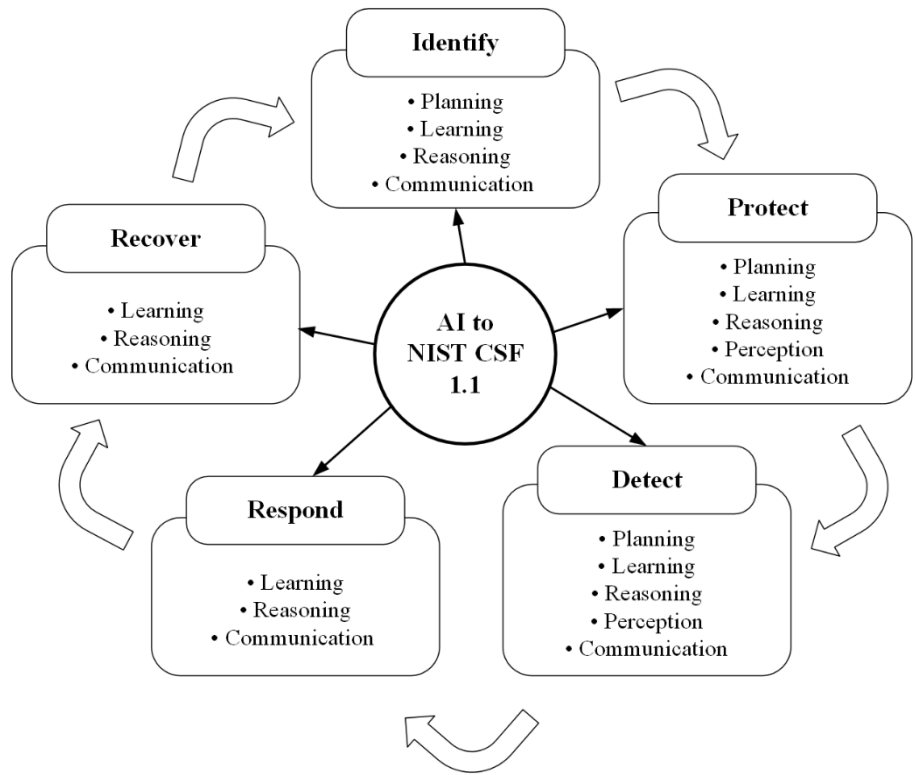

**Fig. 4** AI domain per the NIST CSF 1.1 function: Shows how AI-based techniques map to five original NIST CSF 1.1 functions (Identify, Protect, Detect, Respond, Recover) before the addition of the Govern function

A further limitation in Kaur et al. [101] approach is their reliance on terms like 'AI-based,' 'AI-powered,' and 'AI can automate,' suggesting a focus on automation or assisted intelligence (refer to Fig. 3). This contrasts with the exploration of AI agents (autonomous intelligence), a critical aspect of this paper. The phrase 'AI agent' is used only once in their work. While the AI automation techniques in Fig. 4, such as planning, reasoning and learning, exhibit some overlap with basic AI agent properties discussed in the introduction section and elaborated upon further in Section 2.2.3, the study, like Pleshakova et al. [99] and Hauptman et al. [100], did not address the NIST CSF 2.0. Therefore, the systematic mapping of AI agents to the current NIST CSF 2.0 remains a conspicuous research gap that requires further investigation.

## 2.2 Foundations of AI agents

### 2.2.1 Large-language models

Driven by DL architectures and massive pre-training, LLMs have emerged as a transformative force in natural language processing [114]. With transformer-based architectures like OpenAI's Generative Pre-trained Transformer (GPT) series, Google's Gemini series, and Anthropic's Claude series, LLMs showcase unparalleled abilities in capturing and generating human-like text [92], [115]. However, this general-purpose proficiency masks critical limitations. LLMs, without customisation, struggle with domain-specific knowledge and proprietary data, making them unsuitable for many real-world applications [114]. The prohibitive cost and resource demands of training LLMs from scratch further necessitate the development of tailored solutions [116]. As a result, the field has witnessed an explosion of customisation strategies, broadly falling into two distinct categories, designed to adapt LLMs to specific application contexts:

- *Parameter-efficient fine-tuning (PEFT) or frozen model adaptation:* PEFT techniques adapt pre-trained LLMs by training only a small subset of newly introduced parameters, preserving/freezing the original model's weights and significantly reducing computational costs [116].
- *Full fine-tuning*: In contrast, full fine-tuning modifies all parameters of a pre-trained LLM to optimise performance on a specific task, demanding more computational resources and potentially altering the model's general knowledge [117].

After selecting a foundational LLM, a customisation strategy is required to facilitate the adaptation of the LLM for specialised applications. This paper presents a spectrum of five prevalent customisation strategies, ordered by progressively increasing resource demands and computational expenditure:

- *Prompt engineering (Strategy: PEFT-adjacent):* This strategy, which includes in-context learning and requires minimal resource investment, focuses on carefully crafting input prompts to elicit desired LLM responses, effectively steering the model's inherent capabilities without parameter modification [118].
- *Retrieval augmented generation (RAG) (Strategy: PEFT-adjacent):* Employing moderate resource consumption, RAG augments LLM responses with external knowledge retrieval, enhancing accuracy and relevance by incorporating real-time information [119].
- *Agent frameworks (Strategy: PEFT-adjacent/Hybrid):* These agentic frameworks, with increasingly resource-intensive demands, enable LLMs to interact with external environments and tools, facilitating the execution of complex tasks through dynamic interaction behaviour [71]. This approach can be seen as a hybrid, as some agents may modify small parameters while others do not.
- *Fine-tuning (Strategy: Full fine-tuning):* This strategy involves significant resource allocation to adapt LLM parameters using domain-specific datasets, resulting in enhanced performance for targeted applications [117], [118].
- *Reinforcement learning from human feedback (RLHF) (Strategy: Full fine-tuning/Hybrid):* Requiring maximum resource expenditure, RLHF refines LLM behaviour through human preference-based learning, aligning the model's outputs with desired human values and preferences [119]. This also has hybrid qualities, as some implementations only modify the "reward" model, while others modify the base LLM.

Customisation strategies designated as 'PEFT-adjacent/Hybrid' represent approaches that, while not strictly adhering to standard PEFT methodologies, either share core principles of minimising pre-trained weight modifications or integrate aspects of PEFT and full fine-tuning. Similarly, the designation 'Full fine-tuning/Hybrid' indicates strategies that predominantly rely on full fine-tuning, modifying all model parameters, but may also integrate aspects of parameter-efficient methods or possess implementations that vary, resulting in a combination of full and parameter-efficient fine-tuning characteristics.

This paper focuses on agent frameworks, a customisation strategy that enables LLMs to interact with external environments and tools. On November 25, 2024, Anthropic [120] introduced the Model Context Protocol (MCP), a new standard for connecting AI models to diverse external data systems, including business tools, content repositories, and development environments. Thus, the MCP is an open standard designed to simplify the integration of AI agents with real-world data, providing context to underlying LLMs [121]. Additionally, in collaboration with a wide range of industry partners (Deloitte, Langchain, Salesforce, Cohere, and fifty-six others), Google's launch of the Agent2Agent (A2A) protocol on April 9, 2025, an open standard that complements the MCP, is poised to accelerate the evolution of interconnected AI agents by enabling cross-vendor interoperability between agents [122]. A good understanding of AI agents necessitates thoroughly examining their core principles, inherent properties, and the range of agent types.

### 2.2.2 Defining AI agents

Building on the discussion of LLM customisation strategies, this paper explores the critical role of AI agents within agentic frameworks. To contextualise this, it is essential to clarify the concept of 'agency' in AI systems. 'Agency' denotes the extent to which an AI system's behaviour is goal-directed, directly impacts its environment, and enables it to achieve long-term objectives with minimal human intervention [71]. This implies a shift from passive AI tools to active decision-making entities. 'Agentic AI,' therefore, represents the overarching framework that empowers AI agents to operate with heightened autonomy. It encompasses the broader capabilities that enable LLMs to function as agents, facilitating complex interactions and decision-making processes [123]. In this context, an AI agent is a specific instantiation or component within an agentic AI system designed to execute particular tasks or functions. This distinction between the broader agentic AI framework and the specific AI agent implementation is crucial for understanding the nuanced application of LLMs in autonomous systems. Recognising that agentic AI is a system and an AI agent is a component within that system is very important.

While scholarly definitions of AI agents vary, they converge on the fundamental concept of systems that perceive their environment and autonomously execute actions to optimise goal achievement [25]. These systems span from basic rule-based programs to sophisticated learning and reasoning entities [30]. Despite definitional nuances, the core tenets of AI agents consistently emphasise their capacity for independent and intelligent operation within a given environment [26], [27]. Table 3 summarises AI agent definitions according to several authors.

**Table 3** AI agent definitions: Provides various scholarly definitions of AI agents, capturing themes such as autonomy, adaptability, and environment interaction

| AI agent definition | Author |
| --- | --- |
| A programmed entity that performs operations on behalf of another user or program with a certain degree of independence. | Alrfai et al. [124] |
| Systems capable of executing tasks without human intervention. | Dehghantanha et al. [68] |
| Software or a collection of software that resides and operates on one or more computing devices, perceives its environment, and executes purposeful actions on the environment (and on itself) to achieve the agent's goals. | Kott et al. [36] |
| Entities that can perform certain functions without human intervention, including self-activating, self-sufficiency, and persistent computation. | Ligo et al. [38] |
| An entity that is a mixture of hardware and software and uses sensors to perceive its surroundings and actuators to make changes. | Sharma and Jindal [40] |
| Pieces of software or hardware with a processing unit capable of making wise decisions about their courses of action in uncertain and adverse environments. | Théron and Kott [35] |
| Computer-assisted systems that can generate text, images, audio, or videos. | Baabdullah [95] |
| An entity with self-adaption and intelligence in an environment can understand its environment and control its decision behaviour; it conducts self-regulation and self-learning after it perceives changes in the environment, and it can proactively complete preset tasks. | Cao et al. [107] |
| Systems capable of pursuing complex goals with limited supervision. | Chan et al. [71] |
| Systems or software that can perform tasks or functions independently, without human intervention, in various environments and situations are designed to make decisions and take actions based on their programming and the data they receive from their surroundings. | Cronin [72] |
| Autonomous systems that interact with their environment to achieve specific objectives, often using LLMs to enhance natural language processing capabilities. | Han et al. [92] |
| An intelligent entity that can autonomously execute appropriate and contextually relevant actions based on sensory input, whether in a physical, virtual, or mixed-reality environment. | Huang et al. [114] |
| Systems that perceive their environment and take actions to maximise their chances of achieving their goals. | Rossi [25] |
| Software programs that respond to states and events in their environment, independent from direct instruction by the user or owner, acting on behalf and in the owner's interest. | Bösser [30] |

As evidenced by the diverse definitions presented in Table 3, the absence of a unified conceptualisation of AI agents has resulted in a proliferation of varied architectural and methodological approaches. This lack of consistency directly impacts the design and implementation of agentic systems, necessitating a thorough examination of the field's dominant trends and ongoing projects to understand the resulting complexities in modelling and application [26]. Using the various definitions in Table 3, I asked ChatGPT 4.5 to analyse themes and patterns and construct a single definition. The exact prompt is as follows:

*I have compiled a collection of AI agent definitions from various scholarly research papers. Please analyse the following definitions, identify the recurring themes and patterns, and construct a unified definition that encapsulates the core concepts shared across these definitions. Aim for a definition that is both comprehensive and concise.*

*Definitions: [I inserted my list of AI agent definitions from Table 3 here, each clearly separated.]*

*Specifically, please focus on identifying:*

- *Common elements: What core capabilities or characteristics are consistently mentioned?*
- *Variations: Where do the definitions differ, and what are the key points of divergence?*
- *Essential components: What elements are indispensable for a system to be considered an AI agent?*

*Based on your analysis, provide a unified definition of an AI agent that reflects the essence of these definitions.*

The resulting definition is as follows: *An **AI agent** is an autonomous computational entity—comprised of software or hardware and software integration—that independently perceives its environment, intelligently processes information, makes contextually appropriate decisions, and proactively executes purposeful actions to achieve defined goals, adapting dynamically through self-regulation and learning to changing circumstances.*

Although AI agent properties are explored in detail in the next section, the ChatGPT prompt above also yielded core AI agent characteristics emanating from the studies in Table 3. These are:

- *Autonomy*: Capacity to operate independently.

- *Environmental perception*: The capability to gather and interpret environmental data.
- *Decision-making*: Ability to independently select appropriate actions or behaviours.
- *Goal-oriented action*: Pursuit and achievement of specified objectives through purposeful interactions.
- *Adaptability and learning*: The capability for self-adaptation and learning.

In addition to the above, the following section provides a thorough exploration of key AI agent properties.

### 2.2.3 AI agent properties

AI agents possess several key properties that enable them to interact with their environment and make decisions to achieve specific goals. A literature review reveals several properties, as shown in Table 4.

**Table 4** AI agent properties: Lists the key properties of AI agents (e.g., autonomy, learning) identified through a literature review, with brief descriptions and source references

| Core property | Description | Author |
|---|---|---|
| Autonomy | Operate independently without human intervention. | Phillips-Wren, Leite et al., Bösser, Gu and Li, and Liu et al. [22], [23], [30], [125], [126] |
| Adaptive learning | Learn from past experiences and adapt to new situations. | Rabuzin et al., Leite et al., Abiodun and Khuen, Sethy et al., and Mazumder and Liu [22], [24], [26], [127], [128] |
| Proactive goal pursuit | Take initiative to achieve goals. | Phillips-Wren, Leite et al., Jameel et al., Liu et al., and Pantoja et al. [22], [23], [125], [129], [130] |
| Reactive responsiveness | Respond to environmental changes in real-time. | Phillips-Wren, Leite et al., Jameel et al., Liu et al., and Pantoja et al. [22], [23], [125], [129], [130] |
| Inter-agent communication | Share information and coordinate actions with other agents. | Phillips-Wren, Leite et al., Jameel et al., Liu et al., Pantoja et al., and Sethy et al. [23], [125], [128], [129], [130] |
| Collaborative interaction | Collaborate in MAS to achieve shared goals. | Leng et al., Phillips-Wren, Jameel et al., Liu et al, and Sethy et al. [23], [125], [128], [129], [131] |
| Knowledge representation and reasoning | Store and utilise knowledge for decision-making. | Sennott et al. [132] |
| Ethical reasoning and decision-making | Embed ethical decision-making capabilities into AI systems. | Bösser, Rossi and Mattei, and Gu and Li [30], [126], [133] |
| Moral agency (where applicable) | Ensure actions by agents with significant autonomy and decision-making align with moral codes and societal norms. | Bösser, and Gu and Li [30], [126] |
| Social interaction | Interact with humans and other agents through various interfaces to influence their decisions, confidence, and trust through transparency and reliability. | Nickles et al., Phillips-Wren, Leite et al., Rossi and Mattei, Pitardi and Marriott, Chakraborty, He and Jazizadeh, and Peng et al. [19], [22], [125], [133], [134], [135], [136], [137] |
| Trustworthiness | Build user trust through reliability and transparency. | Pitardi and Marriott, and He and Jazizadeh [135], [137] |
| Domain-specific competence | Applications in various fields, such as smart homes, healthcare, and military operations. | Bösser, Rossi, Preethiya et al., Rashid and Kausik, and Liu et al. [25], [30], [32], [138], [139] |
| Explainability | Provide clear and understandable explanations for actions and decisions. | Chakraborty, Majumdar, and Simran et al. [136], [140], [141] |
| Resilience | Maintain performance and functionality in the face of unexpected inputs, noise, or adversarial attacks. | Cam, Falowo et al., and Rafferty and Macdermott [142], [143], [144] |
| Embodiment (if applicable) | In applicable instances, build a physical presence or virtual representation in an environment, enabling it to interact with the world through sensors and actuators. | Abiodun and Khuen, Weng and Ho, Bovo et al., Preethiya et al., and Liu et al. [31], [32], [94], [127], [139] |
| Long-term memory | Retain and utilise information over extended periods. | Chan et al., Kim and Saad, Chen et al., and Deng et al. [71], [93], [98], [123] |
| Agency transfer/Delegation | Transfer or delegate parts of the agency or decision-making to other agents or humans. | Neff and Nagy, Baird and Maruping, and Candrian and Scherer [65], [145], [146] |

As shown in Table 4, AI agents' functionality is enhanced by various core properties, including autonomy, adaptability, and reasoning [23], [30]. While functionalities such as planning, goal-oriented behaviour, and explicit decision-making are undeniably crucial for AI agent operations [28], [30], [127], my review in Table 4 prioritises the examination of broader, foundational properties that underpin these capabilities. These specific functionalities are often observed as emergent behaviours or direct applications of core attributes like autonomy, adaptive learning, and (knowledge representation and) reasoning. For example, sound reasoning inherently enables informed decision-making [30], [132], and planning is intrinsically linked to goal-oriented actions [29], both of which are facilitated by a high degree of autonomy [41], [147]. Furthermore, domain-specific competence integrates and applies these foundational properties within context. In addition, properties such as explainability, resilience, and long-term memory are becoming increasingly relevant as AI agents are deployed in more complex and critical applications [123], [140], [142]. Therefore, this paper focuses on the core properties—autonomy, adaptive learning, proactive goal pursuit, reasoning, and others detailed in Table 4—to provide a framework for understanding the fundamental principles that govern AI agent behaviour while acknowledging the inherent interconnectedness and hierarchical relationships among these attributes.

However, semantic gaps in communication languages challenge the development of AI agents with these properties, and there is a need for firm ethical boundaries [133], [134]. In other words, ethical considerations like fairness, data privacy and security, and transparency are essential for responsible AI development [128], [129], [136]. As Radanliev et al. [148] articulate, autonomous AI systems must be evaluated not only for their technical performance and alignment with established frameworks, such as the NIST CSF, but also, crucially, for their potential to heighten cyber risks, compromise privacy, or operate beyond adequate human control. These profound concerns fundamentally reshape the discourse, underscoring the imperative for agentic frameworks that inherently embed ethical safeguards and responsible AI principles throughout the entire deployment lifecycle.

The limitations of AI agents extend beyond the absence of a unified definition, current technological constraints, and ethical considerations. Notably, legal and regulatory frameworks present a substantial challenge, particularly concerning the classification of autonomous AI systems. Whether such systems should be granted legal personhood rather than being treated as mere objects is a subject of ongoing debate [149]. However, a thorough examination of these legal implications is beyond the scope of this paper. Building on the discussion of AI agent properties and their inherent limitations, the following section provides an overview of their classification, highlighting the diverse types and their respective applications.

### 2.2.4 AI agent types

A clear differentiation based on their functional and task-specific architectures is required to understand the diverse applications of AI agents. In the same SLR exercise described in Section 2.1.3, one of the themes extracted from the qualitative data was under the category of 'AI agent types.' I asked ChatGPT 4.5 to analyse this 'AI agent type' data, and the exact prompt is as follows:

*You are an expert in artificial intelligence and agent-based systems, tasked with analysing the findings of a systematic literature review (SLR) on AI agent types. I will provide you with a series of notes extracted from the SLR, describing various entities that authors have identified as AI agents. Your task is to:*

1. *Categorise the data into broader, more accepted 'AI agent type' categories. These categories should reflect the core, fundamental distinctions in AI agent design and functionality established by reputable academic and industry sources.*

2. *Identify and describe any subcategories or variations mentioned in the notes. Explain how these subcategories relate to the main categories, highlighting their specific characteristics and differences.*

3. *Acknowledge and clarify any instances where authors may refer to specific implementations, applications, or variations rather than distinct agent types. Explain the nuances of these distinctions.*

4. *Present your analysis in a clear, structured table format. The table should include:*
   - *'Main AI Agent Category'*
   - *'Subcategories/Variations'*
   - *'Description and Nuances'*

5. *Provide a concise summary of the key findings and trends identified in the SLR data.*

*Please ensure your analysis is grounded in established AI agent literature and reflects a comprehensive understanding of the field.*

*[Insert your notes from the SLR here. In my case, I attached a document with my SLR notes.]*

Table 5 presents a structured analysis of the SLR data, organised into AI agent categories, including subcategories and nuances that are clarified.

**Table 5** AI agent types: Classifies AI agent architectures (e.g., reactive, cognitive) and subcategories or variations, showing how each addresses distinct design and operational nuances

| AI agent category | Subcategories/Variations | Description and nuances |
|---|---|---|
| Autonomy-based agents | o Autonomous agents<br>o Adaptive autonomous agents<br>o Continual learning-enabled AI agents (learning agents) | Agents perform tasks independently, with autonomous agents operating without humans [41], [150]. Adaptive autonomous agents dynamically adjust autonomy levels according to contextual requirements [100]. Learning agents continuously self-learn, adapting to novel situations autonomously [23], [24]. |
| Interaction complexity (Agent cognition) | o Simple reflex agents<br>o Model-based reflex agents<br>o Goal-based agents<br>o Utility-based agents<br>o Cognitive agents | Categorised by cognitive complexity, simple reflex agents react directly to current percepts without history. Model-based reflex agents maintain internal states by tracking environmental aspects [151]. Goal-based agents explicitly pursue defined objectives, guiding their actions. Utility-based agents select actions that maximise outcomes' desirability or utility [151]. Cognitive or deliberative agents are intelligent agents that aim to emulate human-like cognitive processes [152]. They are responsible for higher-level tasks such as learning, planning, conflict resolution, and task management [28]. |
| Embodiment | o Embodied agents<br>o Virtual agents | Embodied agents possess physical forms (such as robots and drones) capable of physical actions [31], [32]. Virtual agents exist entirely as software, operating digitally, such as chatbots and virtual assistants [31]. |
| Behavioural approach | o Reactive agents<br>o Proactive (Goal-oriented) agents<br>o Hybrid agents<br>o Learning agents | Reactive agents respond directly to stimuli without using internal states or past experiences [29]. Proactive agents utilise internal states, planning, and past experiences to achieve goals [29]. Hybrid agents incorporate both reactive and proactive functionalities, dynamically adjusting to their surroundings [29]. |
| Collaboration and coordination | o Cooperative agents<br>o Centralised agents<br>o Decentralised agents<br>o Multi-agent systems (Swarm, Cohort, Society of agents)<br>o Autonomous collaborative agents | Cooperative agents collaborate toward common goals in MASs [150], [153]. Centralised agents are centrally coordinated, whereas Decentralised agents coordinate independently through communication [153]. MASs demonstrate collective behaviours through horizontal (swarm) or vertical (cohort) coordination, enabling emergent properties and self-organisation [35], [36], [110]. Autonomous collaborative agents independently perform complete tasks, interacting with others when necessary [35], [36]. |
| Ethics and morality | o Artificial moral agents | Designed explicitly to facilitate ethical decision-making, adhering to moral standards is crucial in ethically sensitive applications such as healthcare or autonomous vehicles [126]. |
| *Architectural design paradigms* | o *Thinking-type architecture (Symbolic/Decision-based)*<br>o *Response-type architecture (Event-driven)*<br>o *Mixed-type architecture (Hybrid)* | *Thinking-type agents execute actions through symbolic reasoning and decision-making processes. Response-type agents primarily react to external perceptions or events. Mixed-type agents combine symbolic reasoning and reactive perception-driven behaviour* [107]. |
| *Agent implementation contexts* | o *Specialised agents (Functional specialisation)*<br>o *Fully functional agents (Holistic capability)* | *Specialised agents focus on performing discrete functional roles as part of broader collaborative agent societies. Fully functional agents independently execute all necessary functions for their designed purpose and are capable of autonomous operation* [35], [36]. |
| *Specific implementations and applications (Contextual examples)* | o *Self-driving cars*<br>o *Drones*<br>o *Robotic vacuum cleaners*<br>o *Chatbots and virtual assistants*<br>o *Text-based generative agents*<br>o *Automated trading systems*<br>o *Robotic space exploration probes*<br>o *Internet information agents* | *Contextual examples demonstrating practical implementations utilising core agent characteristics (autonomy, cognition, embodiment). Examples: autonomous vehicles, drones, home robots, automated financial trading* [72], *robotic space exploration* [40], *conversational AI and generative text-based agents* [72]. *Such examples reflect real-world applications rather than distinct agent types.* |

As evidenced through Table 5, the SLR identifies distinct AI agent categories based on autonomy, cognitive complexity, embodiment, behavioural approach, coordination structure, ethical considerations, and architectural design. Trends indicate a growing emphasis on adaptive autonomy, continual learning capabilities, ethical and moral agency, and multi-agent collaboration for complex problem-solving. Contextual examples provided in the literature, listed in the last row of Table 5, emphasise practical applications rather than new agent categories, highlighting how theoretical constructs of AI agents manifest in varied real-world scenarios. It is clear from Table 5 that not all agent types are equally suited to every security task. The framework proposed in this paper aims to provide a systematic and contextually relevant approach to selecting AI agents suitable for cybersecurity tasks.

## 3. Method for developing an AI agent selection framework

Driven by the need for a structured approach to AI agent selection, design, and deployment, informed by a systematic categorisation of agent types and their characteristics, I present the AI Agent Taxonomy and Decision Framework (AIATDF). This framework, grounded in literature-derived agent properties and categories, provides a decision-making tool for mapping suitable AI agents to the NIST CSF 2.0. The framework development method, guided by the AI agent properties in Table 4, the architecture/types in Table 5, and the NIST CSF 2.0 core structure in Fig. 2, is visually represented in Fig. 5.

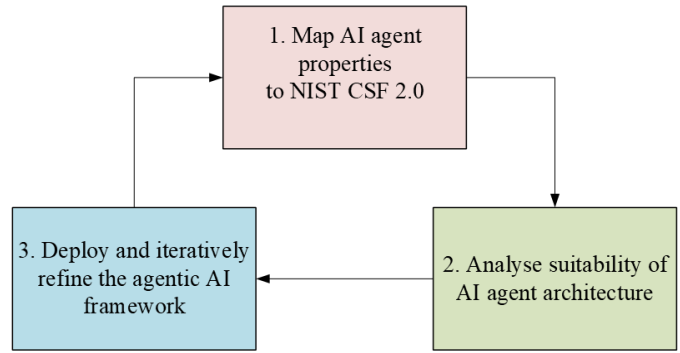

**Fig. 5** AIATDF development method: Outlines the overarching method employed in the study for the development of the AI agent taxonomy and decision framework

Essentially, Fig. 5 says the following:
1) *Establish agent property-NIST CSF 2.0 task alignment:*
    o Map AI agent properties to specific NIST CSF 2.0 task requirements.
    o Method: Develop a comprehensive mapping matrix.
2) *Conduct architectural suitability assessment:*
    o Evaluate the architectural suitability of various AI agent types for NIST CSF 2.0 task requirements.
    o Method: Perform a comparative analysis of agent architectures.
3) *Implement and iteratively refine agent deployment:*
    o Deploy selected agents and establish a continuous refinement process.
    o Method: Implement a feedback-driven iterative optimisation cycle.

Executing the steps above yields the AIATDF presented and explained in the next section.

## 4. AI agent taxonomy and decision framework

In the AIATDF description, 'taxonomy' emphasises the systematic categorisation of agent types derived from the literature, and 'decision framework' indicates a structured approach for guiding the selection, design, and deployment decisions of AI agents based on identified agent categories, variations, characteristics, and application contexts. Following the three framework development steps outlined in the methods section, Fig. 6 represents the AIATDF of this paper.

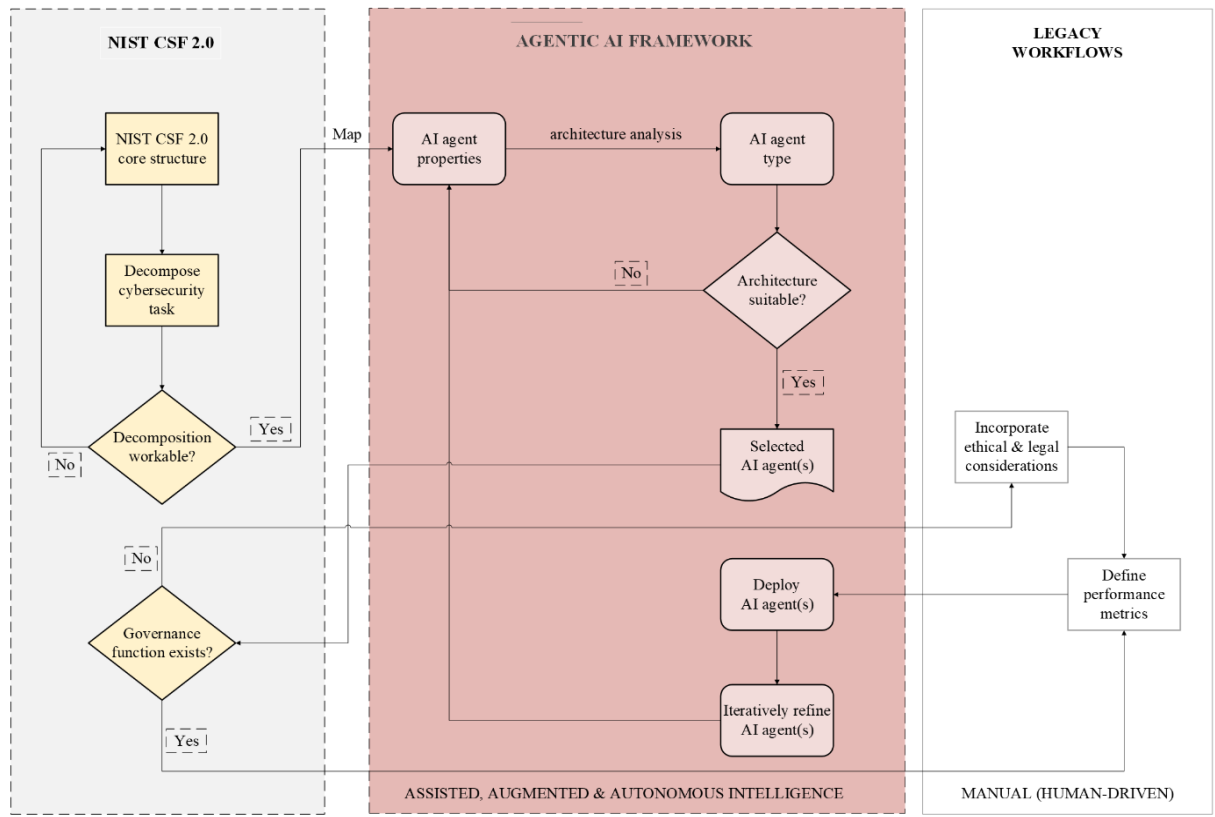

**Fig. 6** AI agent taxonomy and decision support framework: Provides an overview of the six main stages of the proposed AIATDF, showing how each step contributes to choosing and deploying AI agents aligned with the NIST CSF 2.0

The AIATDF comprises six main stages as follows:

1) *Contextual cybersecurity task decomposition:*
    o Break down complex cybersecurity tasks into granular subtasks within the NIST CSF 2.0.
    o *Method*: Hierarchical task analysis of the NIST CSF 2.0's subcategories through the 'informative references' functionality.
2) *Mapping agent properties to NIST CSF 2.0:*
    o Link AI agent properties to the NIST CSF 2.0 cybersecurity task-specific requirements.
    o *Method*: Create a mapping matrix.
3) *AI agent-type architectural suitability analysis:*
    o Assess AI agent architecture suitability for NIST CSF 2.0 cybersecurity task-specific requirements.
    o *Method*: Comparative analysis of agent architectures.
4) *Performance evaluation metrics for agent effectiveness:*
    o Define metrics to assess AI agent(s) performance in NIST CSF 2.0 objectives.
    o *Method*: Use industry best practices to define key performance indicators.
5) *Design, develop, and deploy AI agent(s)*
    o Create and implement AI agent solutions that meet the architectural and performance requirements identified in the previous stages.
    o *Method*: Use established software development practices (e.g., Agile or DevSecOps) and pilot testing to integrate the chosen agent architectures into the organisation's SOC or equivalent cybersecurity workflow.
6) *Iterative framework refinement and validation:*
    o Establish a feedback loop for continuous improvement of the agentic AI framework.
    o *Method*: Collect data and feedback from cybersecurity practitioners and end-users.

In addition to the six main stages of the AIATDF, there are two extra conditional stages in Fig. 6. These are the 'governance decision' and 'ethical and legal considerations' stages. As discussed in the literature review section, the NIST CSF 2.0 has a 'govern' function that ensures the integration of ethical and legal considerations, the 'yes' option of the 'governance decision' box in Fig. 6. The 'no' governance option in the AIATDF demonstrates that any acceptable cybersecurity framework can be utilised in place of the NIST CSF 2.0, and the agentic AI framework will still be applicable. In other words, any widely recognised cybersecurity standards, guidelines, and frameworks can be utilised with the AIATDF. In the next section, I use the NIST CSF 2.0 to validate the AIATDF.

## 5. Validation of the AIATDF

To establish the validity of the AIATDF—specifically, its ability to represent its intended construct accurately [154]—this test focuses on mapping AI agent properties to six core functions of the NIST CSF 2.0 (Fig. 2). Notably, the initial hierarchical task analysis stage of the AIATDF, which decomposes the NIST CSF 2.0 (sub)categories via the framework's 'informative references,' falls beyond the scope of this agentic AI framework (Fig. 6) and is therefore not validated in this paper. Instead, the validation process commences with the crucial step of aligning AI agent properties (Table 4) to the NIST CSF 2.0 functions. As stage 2 of the AIATDF dictates, this alignment must be documented in a mapping matrix, which directly links AI agent characteristics to specific cybersecurity task-specific requirements (NIST CSF 2.0 functions in this validation exercise). Furthermore, the AIATDF validation test recognises that AI agent types (Table 5)—which directly influence the agentic AI framework architecture—are derived from these properties. Therefore, the validation extends to stage 3, which assesses the suitability of AI agent architectures for the NIST CSF 2.0 task-specific cybersecurity requirements, a process also reflected in the mapping matrix.

For demonstrative purposes, the mapping matrix (Table 6) showcases the application of reactive, cognitive, hybrid, and learning AI agent types; however, it is emphasised that this mapping strategy is universally applicable across all AI agent types. In other words, Table 6 validates a simplified version of the AIATDF conceptual framework. Thus, real-world deployments may use MAS [92] for a single cybersecurity function or combine multiple functions within the same agentic system.

**Table 6**. Mapping matrix for selection, design, and deployment of AI agents: Aligns four AI agent types—reactive, cognitive, hybrid, and learning—with the six NIST CSF 2.0 functions, offering a practical guide to optimal agent deployment

|  | **Reactive agents** | **Cognitive agents** | **Hybrid agents** | **Learning agents** |
|---|---|---|---|---|
| **Govern** | o Enforces static policy checks in real-time<br>o Limited scope for policy evolution | o Policy-driven decision-making & and compliance management<br>o Tracks strategic goals, compliance, risk appetite | o Ensures immediate policy adherence and higher-level oversight<br>o Facilitates cross-functional governance tasks | o Learning-based governance optimisation (e.g., balancing privacy vs. security trade-offs)<br>o Automates policy refinement for dynamic changes |
| **Identify** | o Rapid scanning of known assets and threats<br>o Trigger-based checks for new devices or vulnerabilities | o Strategic planning for comprehensive asset inventories<br>o Incorporates organisational risk models and compliance | o Balances immediate scanning with higher-level threat modelling<br>o Layered architecture for real-time triggers and deeper analysis | o Adapts risk profiles over time via continual learning<br>o Discovers unknown vulnerabilities from historical patterns |
| **Protect** | o Immediate enforcement of access rules<br>o Real-time reactive lockdowns under suspicious activity | o Policy-driven configuration management<br>o Plans system-wide changes and resource allocations | o Integrates rules-based triggers with strategic access control<br>o Quick local responses plus deliberative oversight | o Uses anomaly detection to refine protective measures<br>o Learns evolving user access patterns for dynamic policy updates |
| **Detect** | o Real-time intrusion triggers via signature-based checks<br>o High-speed event-driven anomaly alerts | o Considers multi-step reasoning for stealthy threats<br>o Goal-based analysis of suspicious patterns | o Fast detection loop (reactive layer)<br>o Complex threat correlation (deliberative layer) | o Adaptive models refine detection thresholds over time<br>o Identifies zero-day exploits using ML or RL |
| **Respond** | o Automated containment (block IP, isolate assets), minimal strategic foresight but instant action | o Deliberate coordination of multi-step mitigation plans<br>o Accounts for long-term system impacts | o Instant quarantines combined with policy-based orchestration<br>o Layered approach to reduce escalation time | o RL to optimise response effectiveness<br>o Learns from each incident to improve future mitigations |
| **Recover** | o Scripted fallback to restore systems quickly<br>o Lacks deeper analysis of root causes | o Strategic resource allocation for restoration<br>o Coordinated plan to rebuild or reconFig. critical infra | o Rapid initial recovery plus holistic follow-up<br>o Reacts quickly yet integrates deliberative analysis for improvement | o Adaptive post-incident learning (root-cause analysis)<br>o Uses data from each breach to refine future recovery blueprints |

As shown in Table 6, each row corresponds to one of the six NIST CSF 2.0 functions, and each column highlights which and how selected AI agent types can address the cybersecurity task-specific requirements linked to each function. The bullet points summarise the key capabilities or actions of AI agents. The AI agent capabilities are derived from Table 5. For instance, reactive agents rely on stimulus-response mechanisms and thus react to environmental changes in near real-time [29]. Therefore, they would suit any trigger-based or event-driven cybersecurity activities in the NIST CSF 2.0. The cognitive agents are responsible for higher-level functions such as planning, reasoning, learning, and task management [28], [152]. They would be appropriate for NIST CSF 2.0 cybersecurity tasks that require coordination, strategic planning, multi-step reasoning, and decision-making. Hybrid agents combine all types of AI agents, as shown in Table 6. For instance, they combine reactive and cognitive strategies, typically arranged in layered architectures, so quick reactive behaviours can coexist with higher-level reasoning processes [22], [29]. They are, therefore, suitable for cybersecurity activities that require immediate reaction to events and strategic oversight that requires continuous planning, reasoning and decision-making. Lastly, learning agents can enhance performance by revising internal models based on new data or feedback loops [41], [150], making them well-suited for ever-evolving threat landscapes. I discuss the implications of the AIATDF in the next section.

## 6. Discussions

The proposed framework of the study, the AIATDF, was presented in Fig. 6 and conceptually validated in Table 6. The AIATDF demonstrates how AI agents can automate (assisted intelligence), autonomize (autonomous intelligence), or combine automation and autonomy (augmented intelligence) to meet the NIST CSF 2.0 cybersecurity task-specific requirements.

### 6.1 Implications of the proposed framework

#### 6.1.1 Theoretical insights

This paper introduces a framework that aligns core AI agent capabilities, such as autonomy, adaptive learning, proactivity, and reactivity, with the six fundamental functions of the NIST CSF 2.0. This integration establishes a conceptual bridge, translating abstract AI agent theory into actionable cybersecurity strategies defined by the NIST CSF. In contrast to fragmented, single-capability AI deployments, this holistic framework provides a unified perspective, defining how diverse agent types can address the complex demands inherent in each NIST CSF 2.0 function. By mapping AI agent capabilities to the task-specific outcomes of the NIST CSF 2.0, the AIATDF mitigates the risks associated with haphazard AI adoption. Rather than deploying advanced AI tools without a clear integration strategy, this approach fosters an alignment between agent theory and NIST CSF constructs. Additionally, this alignment enables the design of AI systems that excel in threat detection and response, adapt to evolving vulnerabilities, engage in strategic planning, and enhance governance. Combining theoretical concepts, such as autonomy and AI agent cooperation, with practical objectives, like reduced incident response times and enhanced compliance, empowers organisations to invest in AI-driven cybersecurity solutions with confidence. Consequently, the AIATDF enhances the theoretical rigour and real-world applicability of agent-based defence strategies, ensuring alignment with recognised cybersecurity standards.

One of the key contributions of this study is that the AIATDF also distinguishes between assisted, autonomous, and augmented modes of AI systems intelligence. This contrasts with traditional agent taxonomies that obscure critical distinctions in agent autonomy through broad, overlapping categories. The AIATDF illuminates the nuanced spectrum of AI agent capabilities by explicitly mapping each intelligence mode to specific cybersecurity functions—ranging from rapid, automated tasks (assisted intelligence) to complete human-out-of-the-loop decision-making (autonomous intelligence). In other words, the AIATDF highlights the graduated spectrum of AI agent capabilities. This clarity is crucial; for instance, it describes why a reactive AI agent may be sufficient for immediate, rule-based detection tasks within the Identify or Protect functions, while a learning or cognitive AI agent is indispensable for strategic governance or comprehensive vulnerability assessments. Consequently, the AIATDF demonstrates that "autonomy" is not a monolithic concept but a dynamic attribute that must be tailored to the specific demands of each security context. This granular approach underscores the importance of aligning AI agent architectures with functional requirements. The AIATDF advances the theoretical understanding of agentic AI operations in cybersecurity by emphasising these roles. It reveals how diverse AI agent architectures can be synergistically deployed to achieve operational efficiency and adaptive resilience, moving beyond simplistic classifications to a more nuanced, context-aware deployment strategy.

Furthermore, the AIATDF distinguishes itself by treating AI agent architecture and operational modes as intrinsically interdependent dimensions, thereby providing a unified analytical lens for evaluating and deploying AI agents. Traditional approaches often treat these aspects in isolation, focusing either on architectural sophistication or the desired level of automation, neglecting their critical interplay. In contrast, the AIATDF demonstrates that an AI agent's architectural complexity directly dictates its potential for autonomy, and conversely, the intended autonomy influences the required architectural design. For example, while learning agents can unlock advanced autonomous capabilities, they may still operate in assisted or augmented modes when strategic human oversight is essential. This multidimensional perspective compels researchers to transcend simplistic, single-axis classifications, such as "low" versus "high" autonomy, and examine how an AI agent's structural design and functional objectives converge to address specific cybersecurity demands. This integrated approach facilitates the development of more sophisticated theoretical models, situating layered AI agent architectures within established organisational risk management standards, such as the NIST CSF. Consequently, the AIATDF fosters a holistic understanding that connects context-aware AI agent behaviours with the hierarchical needs of enterprise security, spanning real-time defence to strategic governance. This unified perspective bridges the gap between AI agent design and operational deployment, enhancing theoretical and practical applications.

While AI agents promise significant operational efficiency and scalability in cybersecurity, their deployment introduces novel cyber risks, especially within critical infrastructure domains. As Radanliev et al. [148] and Radanliev [76] emphasise, the rapid integration of AI into critical systems necessitates a thorough examination of both ethical risks and their associated governance structures. This study recognises that autonomous and semi-autonomous AI agents, particularly those interacting with sensitive data streams, can amplify adversarial threats such as model inversion attacks, data poisoning, and adversarial learning. Such vulnerabilities jeopardise not only system integrity but also intellectual property and classified information. Consequently, the AIATDF must explicitly integrate principles of responsible AI deployment to pre-empt unintended harms. Radanliev [76] and Radanliev et al. [148] advocate for ethical frameworks that prioritise transparency, human oversight, contextual accountability, and resilience against manipulation—considerations that become acutely critical in high-stakes environments, such as biomedical research, national security operations, and AI-driven policymaking, where the consequences of compromise are profound. Therefore, my proposed maturity-based framework necessitates supplementation with institutional risk assessments, comprehensive model audit trails, and thorough ethical readiness evaluations prior to full-scale deployment. Future enhancements to the AIATDF should embed ethical assurance layers, including XAI protocols, bias detection modules, and accountability matrices, directly into agentic system design, fostering not just technical adequacy but also societal trustworthiness.

### 6.1.2 Practical significance

The AIATDF offers a structured, step-by-step methodology that enables the mapping of AI agent capabilities to the subcategories of the NIST CSF 2.0, thereby empowering SOC managers and cybersecurity architects with a valuable tool. It eliminates the reliance on inefficient trial-and-error approaches or generic AI solutions, allowing practitioners to systematically identify the optimal AI agent type for each subcategory's operational objectives. For example, reactive agents can be strategically deployed for high-volume, rapid-response tasks, such as real-time anomaly detection. In contrast, cognitive agents may be reserved for strategic planning and governance functions. This targeted approach minimises resource waste and efficiently allocates computational infrastructure and staff training. By providing clear guidance, the AIATDF accelerates the deployment of effective AI-driven solutions, optimising their impact by ensuring an alignment between AI agent strengths and specific cybersecurity requirements. Open standards, such as the MCP, can further streamline these deployments by simplifying the integration of LLM-driven agents with diverse enterprise tools and data repositories, thereby expanding the AIATDF's practical reach.

Furthermore, the AIATDF offers scalability, empowering organisations across varying cybersecurity maturity levels to adopt AI-driven capabilities incrementally. The framework facilitates a strategic entry point for resource-constrained teams through assisted intelligence, focusing on automated alerts, anomaly detection, and recommendation systems that enhance existing human oversight. As organisational confidence and resource allocation expand, the AIATDF supports transitioning to more advanced autonomous solutions, including AI agent-driven incident response and strategic threat analysis. This phased approach minimises operational disruption and fosters stakeholder buy-in, ensuring that each increment of autonomy is implemented optimally for maximum impact. The framework also provides a roadmap for continuous advancement, guiding organisations from foundational automation to sophisticated, fully autonomous AI agent deployments. This strategy enables organisations to adapt and evolve their cybersecurity posture in response to the ever-changing threat landscape, ensuring sustained resilience and operational efficiency.

## 6.2 Comparisons with prior work

It was discussed in Section 2.1.3, and shown in Table 2, that only three studies from prior works mapped AI solutions to the NIST CSF. These studies were conducted by Pleshakova et al. [99], Hauptman et al. [100], and Kaur et al. [101]. This paper distinguishes itself from prior research by presenting a detailed, NIST CSF 2.0-aligned framework for the deployment of AI agents in cybersecurity. Unlike earlier works that focused on limited NIST CSF 1.1 functions [99] or explored AI-based methods without systematic mapping, the AIATDF explicitly integrates a broad spectrum of AI agent architectures with all six NIST CSF 2.0 functions, including the critical 'Govern' function. This approach addresses the fragmented scope observed in studies that concentrated on narrower functions or outdated frameworks, and it builds upon research recognising AI's importance in risk management [101] by formalising the mapping of AI agent taxonomies to security tasks. Furthermore, while some studies explored limited AI autonomy within specific NIST CSF functions [100], the AIATDF spans the entire cybersecurity lifecycle, clarifying how varying degrees of agent autonomy—assisted, augmented, and fully autonomous—intersect with NIST CSF 2.0 subcategories. By systematically linking AI agent properties to these subcategories, the AIATDF provides a promising, theory-driven mapping mechanism, a feature notably absent in prior partial alignments. Thus, the AIATDF bridges previously disparate strands of AI agent research, MAS approaches, and the evolving NIST CSF 2.0, offering a unified mapping framework that extends and reinforces the applicability of AI in modern cybersecurity practices.

## 6.3 Limitations of the study

Although the mapping framework is rigorously grounded in peer-reviewed AI agent research and the established NIST CSF 2.0 guidelines, it remains a theoretical construct. While this conceptual foundation provides a strong analytical framework, the AIATDF *lacks empirical validation* within realistic cybersecurity environments. Consequently, critical questions regarding practical performance—such as false positive rates in anomaly detection or mean time to respond during active attacks—remain unanswered. To bridge this gap, pilot studies or proof-of-concept deployments are essential. These empirical investigations would provide invaluable insights, verifying whether each AI agent type delivers the predicted benefits under real-world operational constraints, including limited computational resources and dynamic threat vectors. This real-world validation is crucial for refining the framework, ensuring its applicability and accuracy in reflecting the complexities inherent in cybersecurity operations.

Although the AIATDF provides distinct AI agent categories for analytical clarity, real-world implementations rarely adhere to such *rigid classifications*. AI agent architectures often exhibit fluidity, particularly in hybrid or evolving systems. For instance, an initially reactive agent may progressively integrate learning components, or a cognitive agent may acquire adaptive capabilities over time. Consequently, an organisation's deployment may deviate from these idealised categorisations, reflecting a more dynamic and integrated expression of AI agent behaviours. This inherent overlap does not invalidate the AIATDF's utility. Instead, it underscores the necessity for context-sensitive applications, emphasising that effective cybersecurity solutions frequently necessitate tailored combinations of AI agent functionalities rather than strict adherence to singular agent types. This context-sensitive approach acknowledges the dynamic and complex nature of real-world cybersecurity challenges, demanding a flexible and adaptable integration of AI capabilities.

The rapid convergence of AI innovation and the evolution of sophisticated threat actors necessitate a dynamic approach to this mapping. Specifically, current AI *agent properties may become obsolete* or superseded by emerging paradigms, such as quantum-safe AI algorithms or next-generation RL. Therefore, the AIATDF must be periodically reviewed and updated to maintain its operational relevance. This proactive adaptation should incorporate novel attack vectors, advanced computational models, and evolving industry best practices. Failure to do so risks deploying AI agents that, while theoretically sound in the current context, may prove inadequate in addressing the dynamic and unpredictable nature of future cyber threats. This continuous evolution is crucial to ensuring that AI-driven cybersecurity solutions remain effective and resilient in the face of emerging challenges.

While the AIATDF provides a promising technical blueprint for aligning AI agent capabilities with specific cybersecurity tasks, it does not explicitly address *critical organisational factors* that influence the sustainable success of AI-driven initiatives. These could be placed under 'legacy workflows' in Fig. 6, and their impact on the AI agent deployment studied. For example, these factors may include organisational culture, budgetary constraints, and staff readiness. In this regard, even a perfectly aligned learning AI agent may fail to deliver its intended value without organisational buy-in, adequate training budgets, or a workforce ready to collaborate effectively with autonomous systems. Consequently, organisations must conduct readiness assessments and develop change management plans to foster internal acceptance and ensure staff possess the requisite skills and tools for seamless integration with AI agents. Without these preparatory measures, even the most technically sophisticated AI deployments risk underutilisation, resistance, or premature abandonment. This underscores the imperative that technological excellence must be coupled with strong organisational alignment to achieve optimal and sustainable outcomes.

## 7. Conclusion

This paper aimed to develop a framework for selecting, designing, and deploying AI agents to address the NIST CSF 2.0 cybersecurity requirements, based on identified agent categories, properties, and application contexts. The study introduced an AI agent taxonomy and decision framework to address this research aim. The AIATDF systematically aligns AI agent types, including reactive, cognitive, hybrid, and learning, with the NIST CSF 2.0 functions (Identify, Protect, Detect, Respond, Recover, Govern), categories, and subcategories.

### 7.1 Summary of key findings

This paper's key findings are encapsulated within five distinct perspectives: (i) theoretical synthesis, (ii) a practical mapping instrument, (iii) a phased adoption strategy, (iv) versatile agent architectures, and (v) framework limitations. Firstly, the AIATDF achieves a novel theoretical synthesis by integrating agent-based AI theory with the structured cybersecurity outcomes defined by the NIST CSF 2.0. This integration, primarily through the differentiation of assisted, autonomous, and augmented intelligence modes, enhances the theoretical understanding of AI agent autonomy's application in cybersecurity. Secondly, the AIATDF functions as a practical mapping instrument, utilising a matrix to demonstrate the suitability of specific AI agent types for the NIST CSF 2.0. This provides SOC managers and cybersecurity architects with a structured, risk-mitigating guide for selecting AI agents. Thirdly, the AIATDF supports a phased adoption approach, enabling organisations to incrementally integrate AI solutions, from basic automated alerts to advanced autonomous and hybrid systems. Fourthly, as detailed in Section 5, conceptual validation demonstrates the AIATDF's comprehensive scope, encompassing all six NIST CSF 2.0 functions and highlighting the potential of a MAS architecture in addressing diverse cyber threats. Finally, while the AIATDF presents a promising theoretical foundation, it requires empirical validation and further development to incorporate socio-technical factors and the evolving capabilities of AI. These five perspectives highlight the potential of a theoretically grounded and practically applicable approach to deploying cybersecurity AI agents. Additionally, they offer academic insight and actionable guidance for enhancing AI-driven defences within the NIST CSF 2.0 framework.

### 7.2 Implications for future research

This study opens several promising research avenues. Firstly, empirical validation and field studies are necessary to establish the applicability of the AIATDF. Pilot deployments in operational environments, measuring metrics such as response times, false positive/negative rates, and resource overhead across diverse organisational contexts (e.g., small vs. large enterprises), would provide essential data for model refinement. Secondly, exploring adaptive and evolving AI agent architectures is imperative, given the rapid evolution of cyber threats and AI technologies. For example, emerging open standards such as the MCP could significantly improve the integration of external data sources into AI agent architectures, rendering the AIATDF approach more scalable and easier to deploy in diverse organisational environments. Therefore, future research should investigate how MAS architectures with adaptive learning layers, leveraging techniques such as reinforcement, transfer, or federated learning, can maintain their efficacy in dynamic threat landscapes and distributed data scenarios. Thirdly, future research should integrate socio-technical factors into AI agent deployment methodologies. Explicitly factoring in organisational culture, budget constraints, and workforce skill sets would facilitate holistic readiness assessments and deepen the understanding of seamless, autonomous, or augmented intelligence implementation. Fourthly, developing ethical and regulatory frameworks is necessary to address the escalating ethical and compliance concerns associated with advanced AI agents. Studies could define guidelines for operationalising responsible AI practices within the NIST CSF 2.0, particularly in high-stakes scenarios. Fifthly, despite the emergence of AI agents CMMs from grey literature, there are no scholarly, well-defined CMMs for adopting AI agents. Future research could provide a structured path (CMM) outlining key stages of AI agent adoption progression and actionable steps for advancement. Ultimately, longitudinal studies examining the long-term deployment of AI agents would provide valuable insights into the evolution of agent-based solutions. These studies would inform continuous improvement mechanisms, ensuring the AIATDF's sustained relevance in the face of evolving AI capabilities and cybersecurity threats.

## Statements and Declarations:

**Ethical Approval and Consent to Participate**

Not applicable.

**Consent for Publication**

The author declares consent for publication.

**Competing Interests**

The author has no relevant financial or non-financial interests to declare.

**Funding**

No funding was received to conduct this study.

**Availability of Supporting Data**

No datasets were generated or analysed during the current study. Only textual data from a systematic literature review underpin the study.

**Author's Contributions**

Conceptualisation, methodology, writing—original draft preparation, and writing—final draft review and editing.

**Acknowledgements**

The author would like to acknowledge the current 'read and publish' agreements negotiated by the South African National Library and Information Consortium (SANLiC) as having contributed to the open-access publication of this paper without the author facing any charges.

**Authors' Information**

Masike Malatji

Graduate School of Business Leadership (SBL), University of South Africa (UNISA), Midrand, Johannesburg, South Africa

**Declaration of Generative AI and AI-assisted Technologies in the Writing Process**

While preparing this work, the author used Scopus AI (Beta) to review some articles, ChatGPT 4.5 to analyse textual data, Google Gemini and ChatGPT 4.5 to structure the initial ideas and logical flow of the paper and improve its readability, and Grammarly for English language editing. After using these AI-powered tools, the author reviewed and edited the content as needed and took full responsibility for the publication's content.